%% file: iccv2017_icpba_ham.tex
\documentclass[10pt,twocolumn,letterpaper]{article}

\usepackage{iccv}
\usepackage{times}
\usepackage{epsfig}
\usepackage{graphicx}
\usepackage{amsmath}
\usepackage{amssymb}

\usepackage{bm}
\usepackage{caption}
\usepackage{xcolor}
\usepackage{relsize}
\usepackage{subcaption}

\newcommand{\T}{\intercal}
\newcommand{\qsection}[1]{\vspace{1.5ex} \noindent \textbf{#1}}

\newcommand{\Mmat}{\mathbf{M}}
\newcommand{\param}{\mathbf{p}}
\newcommand{\paramO}{\mathbf{p}^{(0)}}
\newcommand{\pose}{\bm{\theta}}
\newcommand{\W}{\mathcal{W}}
\newcommand{\pt}{\mathbf{x}}
\newcommand{\ptt}{\mathbf{y}}

\usepackage[pagebackref=true,breaklinks=true,letterpaper=true,colorlinks,bookmarks=false]{hyperref}

\iccvfinalcopy 

\def\iccvPaperID{} 

\ificcvfinal\fi
\begin{document}

\title{Proxy Templates for Inverse Compositional Photometric Bundle Adjustment}

\author{%
\begin{tabular}[t]{c@{\extracolsep{4em}}c}
   \multicolumn{2}{c}{Christopher Ham$^1$, Simon Lucey$^2$, and Surya Singh$^1$}\\
   \rule{0pt}{3ex}
   $^1$Robotics Design Lab                       & $^2$Robotics Institute \\ 
   The University of Queensland, Australia   & Carnegie Mellon University, USA \\
   {\tt\small \{c.ham,spns\}@uq.edu.au}      & {\tt\small slucey@cs.cmu.edu}
\end{tabular}
}


\maketitle


\begin{abstract}
  Recent advances in 3D vision have demonstrated the strengths of photometric bundle adjustment. By directly minimizing reprojected pixel errors, instead of geometric reprojection errors, such methods can achieve sub-pixel alignment accuracy in both high and low textured regions.
  Typically, these problems are solved using a forwards compositional Lucas-Kanade~\cite{lucas1981iterative} formulation parameterized by 6-DoF rigid camera poses and a depth per point in the structure. For large problems the most CPU-intensive component of the pipeline is the creation and factorization of the Hessian matrix at each iteration. For many warps, the inverse compositional formulation can offer significant speed-ups since the Hessian need only be inverted once. In this paper, we show that an ordinary inverse compositional formulation does not work for warps of this type of parameterization due to ill-conditioning of its partial derivatives. However, we show that it is possible to overcome this limitation by introducing the concept of a proxy template image.
  We show an order of magnitude improvement in speed, with little effect on quality, going from forwards to inverse compositional in our own photometric bundle adjustment method designed for object-centric structure from motion. This means less processing time for large systems or denser reconstructions under the same real-time constraints. We additionally show that this theory can be readily applied to existing methods by integrating it with the recently released Direct Sparse Odometry~\cite{engel2016direct} SLAM algorithm.
\end{abstract}


\begin{figure}[h]
\begin{center}
   \newcommand{\warpR}{$\frac{\partial \W(\pt; \param)}{\partial \pose_R} \ne 0$}
   \newcommand{\warpt}{$\frac{\partial \W(\pt; \param)}{\partial \pose_t} \ne 0$}
   \newcommand{\warpd}{$\textcolor{red}{\frac{\partial \W(\pt; \param)}{\partial d} = 0}$}
   \def\svgwidth{0.45\textwidth}
   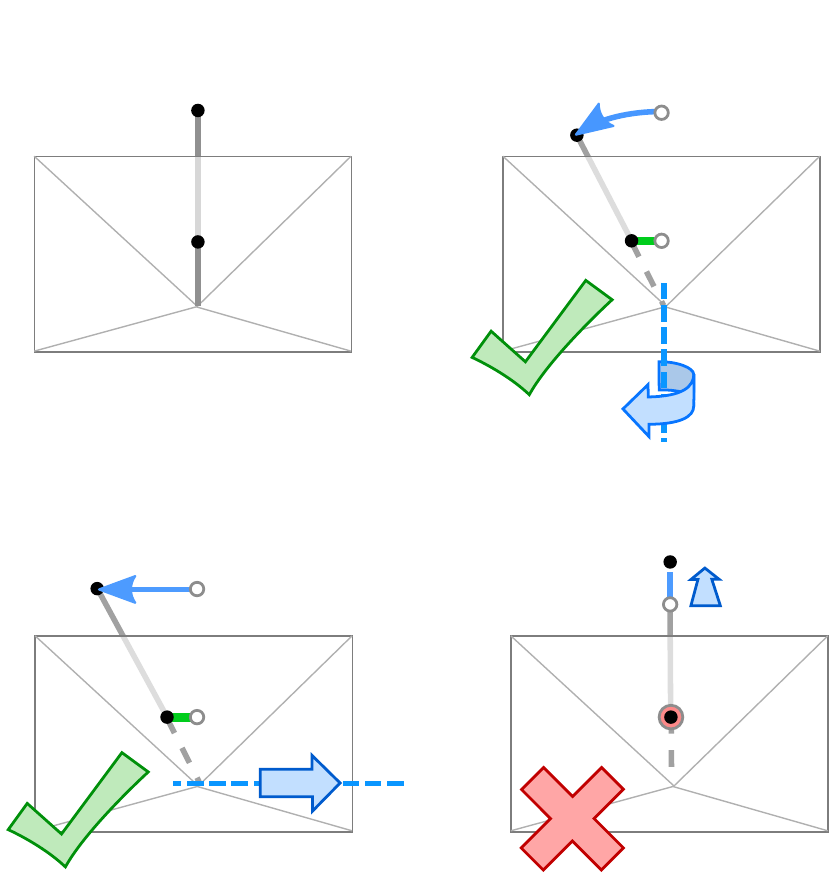
\end{center}
   \caption{The projective function $\W(\pt; \param)$ warps a point into a frame given the point's depth $d$ and frame's pose $\pose$ such that $\param = [ \pose^\T ~~ d]$. The inverse compositional formulation requires that at the identity ($\W(\pt; \paramO) = \pt$)\protect\footnotemark ~all partial derivatives with respect to the parameters be non-zero. We show that this is not the case for depth, and propose an alternative warp that can be used to obtain valid gradients for calculating inverse compositional updates, allowing for efficient large scale photometric bundle adjustment.}
\label{fig:intro}
\end{figure}

\footnotetext{Note that traditionally it is assumed that~$\paramO = \mathbf{0}$ at the identity warp, but as we shall discuss we must be more general when the warping function is by a projected 6 DoF pose and point depth.}

\section{Introduction}
  We are at the cusp of a new wave of reconstruction algorithms that have photometric bundle adjustment (PBA) at their heart. 
  PBA allows for the joint solution of pose and structure with sub-pixel alignment of both high and low textured regions. This is in contrast to feature point bundle adjustment methods which are limited to the precision of their feature detector. Indeed, the field has enjoyed a recent resurgence in direct methods with applications to simultaneous localization and mapping (SLAM)~\cite{engel2014lsd,engel2016direct,alismail2016photometric} and structure from motion (SfM)~\cite{delaunoy2014photometric}. This is thanks to advances in hardware and improved models that have lifted many concerns about computational requirements and robustness.

  That is not to say that such concerns have been eliminated. Current methods use a forwards compositional, gradient descent to jointly solve for 6-DoF camera poses and point locations. While this is sufficiently fast for SLAM problems where the number of cameras and points stays small, the creation and inversion of the Hessian at each iteration becomes slow for larger SfM systems. Typically this is where one might turn to an inverse compositional (IC) formulation so that the Hessian only needs to be generated once. However, we show that ordinary IC cannot be applied directly to systems that parameterize the structure by depths in a reference frame (or template) since the partial derivative of the warping function with respect to the depth is zero at the identity warp (see Fig.\ref{fig:intro}). Further intuition as to why this is the case can be found in considering that \text{any} depth can be chosen for $W(\pt; \param)$ to be the identity warp when there is no rotation or translation.

  A point with inverse depth $d$ can be projected into a target frame with 6 DoF parameters~$\boldsymbol{\theta} \in \mathfrak{se}(3)$ , 
  \begin{align}
    \W(\pt; \param) &= \left< \mathbf{R} \tilde{\pt} + d \mathbf{t} \right> ,
  \end{align}
  where $\left< \cdot \right>$ denotes the projection of a 3D point onto the image plane at $z=1$, and $\param$ is a compact representation of the pose and inverse depth parameters.
  The rotation matrix~$\mathbf{R}$ and translation vector~$\mathbf{t}$ are related to the 6 DoF parameters by the exponential map, 
\begin{align}
    \mathbf{T} =
    \begin{bmatrix}
    \mathbf{R} & \mathbf{t} \\
    \mathbf{0} & 1
    \end{bmatrix} \in \text{SE}(3)
\end{align}
where elements are mapped to $\text{SE}(3)$ by the exponential map~$\mathbf{T} = \exp_{\mathfrak{se}(3)}(\pose)$. 

  The theory of this paper concerns itself with the derivation of the proxy warp $\phi$ used \textit{only} to estimate valid gradients for the inverse compositional Jacobian, and to estimate the updated warp. In its simplest form, we define the warp as
  \begin{align}
    \phi(\pt; \Delta \param) &= \left< \Mmat \left( \mathbf{R} \tilde{\pt} + d \mathbf{t} \right) \right> \label{eq:phi}
  \end{align}
  where $\mathbf{R} = \Delta \mathbf{R} \mathbf{R}_0$, $\mathbf{t} = \mathbf{t}_0 + \Delta \mathbf{t}$ and $d = d_0 + \Delta d$, such that $\phi(\pt; \bm{0}) = \pt$ is the identity warp.

  This alternative warp emerges from the simple idea that instead of taking gradients on the original template, one could first warp the template according to the initialized parameters $\paramO$. The warp that projects a point $\ptt$ from this new template into the target frame is composed as
  \begin{align}
  \W'(\ptt; \param) = \W( \W(\ptt; \paramO)^{-1}; \param)
  \end{align}
  such that $\W'(\ptt; \paramO) = \ptt$ is the identity warp. So long as the origin of the new template no longer coincides with that of the original, we show that partial derivatives are well defined for all parameters.

  In order to estimate residual gradients, we would have to generate a new template for each point-frame combination. To avoid this we warp the point back into the original template according to the initialized parameters and estimate the equivalent gradients. In this way, our new templates become ``proxy'' templates.


  To our knowledge we are the first to propose a method for applying the inverse compositional formulation to photometric systems parameterized by 6-DoF camera poses and point depths, allowing for efficient solutions to large systems. We show experimentally that our method can provide drastic speed-ups with little effect on the accuracy of the result.



\section{Prior Art}  

  Direct methods are often criticized for being susceptible to changes in lighting and the intrinsic photometric response of cameras. While many feature detectors allow one to forget about these concerns, direct methods require that these be explicitly modelled. Far from being a weakness this can be a strength; Engel et al.~\cite{engel2016direct} show that explicit modelling of photometric intrinsics allows their method to remain robust to larger intensity variations than ORB-SLAM~\cite{mur2015orb} (a state of the art feature-based SLAM algorithm).

  The idea of photometric bundle adjustment is young, and the related body of research is small. Other notable methods that formulate a photometric bundle adjustment are from Alismail et al.~\cite{alismail2016photometric}, and Delaunoy and Pollefeys~\cite{delaunoy2014photometric}. These methods choose to parameterize the structure without explicit depths. A closer inspection of their approaches helps motivate why a depth parameterization is a good fit for inverse compositional.

  Alismail et al.~\cite{alismail2016photometric} parameterize the structure by the 3D location of its points, which forces them to ``detach'' each point from its reference frame. A patch of fixed appearance is generated from the reference image for each point. However, detached from the reference frame it becomes challenging to \textit{properly} transform pixels in the neighborhood of the patch center to other frames. Implicitly, the authors approximate this by applying the patch offset after projection. This becomes a poor approximation for large changes in pose (particularly in rotation around the camera's Z-axis).

  Delaunoy and Pollefeys~\cite{delaunoy2014photometric} propose a method for refining camera intrinsics and extrinsics, and object structure parameterized as a watertight mesh. Their variational approach is one of the method's greatest strengths but requires the adandonment of the idea of an explicit reference or template.

  Although they are not photometric bundle adjustment algorithms, both LSD-SLAM~\cite{engel2014lsd} and SVO SLAM~\cite{schops2014semi} show significant computational savings by using pixels directly for short-term pose tracking. In these implementations, the point depths are fixed. With good outlier detection, the direct components of these algorithms offer sub-pixel refinement of the pose parameters without the need to calculate expensive features and search for correspondences. Further, SVO calculates the pose updates using the inverse compositional formulation.

  While photometric bundle adjustment is a relatively new area of research in the computer vision community, the inverse compositional algorithm has been known for some time. The seminal work by Baker and Matthews~\cite{baker2001equivalence} generalized the inverse compositional method to many warps. They proved its equivalence to the forwards algorithm according to a first order approximation, and empirically demonstrated similar performance forwards compositional method when applied to flexible appearance models.

\section{Inverse Compositional for SfM} \label{sec:method}
  Our method allows us to formulate inverse compositional (IC) updates in order to efficiently solve large photometric bundle adjustments. A popular representation of scene geometry is by the inverse depths of 3D points in a reference frame as it accurately models the world while minimizing the degrees of freedom of the structure. While such a representation maintains the concept of a template, we show that it is incompatible with an ordinary IC approach. However, by introducing the concept of proxy templates, we derive a alternative warp that can be used to estimate valid gradients and updates to the original warp.

  Concretely, we are concerned with minimizing the energy
  \begin{align}
    E_{total} &= \sum_f^F \sum_n^N E_{fn} \\
    E_{fn} &= \sum_{\pt \in \mathcal{P}_n} \| I_f(\W(\pt; \param_{fn})) - I_0(\pt) \|_\gamma \label{eq:energy}
  \end{align}
  where $I_0$ and $I_f$ are the reference and target images respectively, $\mathcal{P}_n$ is the set of pixels from the template used for point $n$, $\|\cdot\|_\gamma$ is the Huber norm, and project pixel $\pt$ into frame $f$ by
  \begin{align}
    \W(\pt; \param_{fn}) &= \left< \mathbf{R}_f \tilde{\pt} + d_n \mathbf{t}_f \right>. \label{eq:warp}
  \end{align}

\qsection{Notation.}
  Bold lower-case letters ($\mathbf{t}$) denote vectors, bold upper-case letters ($\mathbf{R}$) denote matrices. Functions over images ($I(\cdot)$) are bilinear interpolations of the underlying image.
  A point $\pt$ is made homogeneous by
  \begin{align}
    \tilde{\pt} = \left[ \pt^\T ~~ 1 \right]^\T.
  \end{align}
  A 3D point is projected into the the image plane by
  \begin{align}
    \left< [ x~~ y~~ z ]^\T \right> = \left[ \tfrac{x}{z}~~ \tfrac{y}{z} \right]^\T.
  \end{align}
  We use $\param_{fn} \in \mathfrak{se}(3) \times \mathbb{R}$ as a compact representation of the 3D rigid camera pose at frame $f$ and inverse depth for point $n$
  where with a slight abuse of notation we assign the first 6 elements as the Lie-algebra representation of the pose $\pose_f \in \mathfrak{se}(3)$ and the final element as the inverse depth $d_n$. 
  \begin{align}
    \param_{fn} =& \left[ \pose_f^\T ~~ d_n \right]^\T \\
    \begin{bmatrix}
    \mathbf{R}_f & \mathbf{t}_f \\
    \mathbf{0} & 1
    \end{bmatrix}
    :=& \exp(\pose_f) \in \text{SE}(3)
  \end{align}
  We define the operator $\boxplus$ such that it applies a left multiplicative update to the pose and additional update to the inverse depth
  \begin{align}
    \Delta \param \boxplus \param = \left[ (\Delta \pose \cdot \pose)^\T ~~ (d + \Delta d) \right]^\T
  \end{align}
  In the following sections we will drop the $f$ and $n$ subscripts for succinct notation unless otherwise required for disambiguation.

\subsection{Ordinary Inverse Compositional} \label{ssec:ic_ord}
  We show that an \textit{ordinary} IC formulation for PBA is not possible since the derivative with respect to the inverse depth is zero at the identity warp, and thus no meaningful gradients can be obtained when calculating the Jacobian.
  Given the warp in (\ref{eq:warp}), we observe that it is the identity warp for $\mathbf{R} = \mathbf{I}$, $\mathbf{t} = \bm{0}$ and any inverse depth
  \begin{align}
  \W(\pt; \paramO) &= \left< \mathbf{I} \tilde{\pt} + d \bm{0} \right> = \left< \tilde{\pt} \right> = \pt
  \end{align}

  By first taking the derivative of the transformed point with respect to the inverse depth before it is projected into the image
  \begin{align}
  \frac{\partial}{\partial d} \left( \mathbf{R} \tilde{\pt} + d \mathbf{t} \right) \bigg\rvert_{\param=\paramO}
  &= \mathbf{t}_0 = \bm{0}
  \end{align}
  we do not even have to apply the quotient rule to see that $\frac{\partial \W}{\partial d} = \bm{0}$.

\subsection{Proxy Templates}
  Instead of using the original reference image $I_0$, let us now consider a warped version $I'_0$ according to a good initialization of the the parameters $\paramO$ (already required by existing PBA methods). Then (\ref{eq:energy}) can be expressed alternatively as
  \begin{align}
    \sum_{\ptt \in \mathcal{Q}} \| I(\W'(\ptt; \param)) - I'_0(\ptt) \|_\gamma \label{eq:newtemp}
  \end{align}
  where the new template point $\ptt = \W(\pt; \paramO)$ is constant, and the new warp $W'$ is composed by the inverse warp of the initial parameters and the original warp such that
  \begin{align}
    \W'(\ptt; \param) &= \W( \W(\ptt; \paramO)^{-1} ; \param) = \W(\pt; \param).
  \end{align}
  At first glance there may be little to gain from this. Note, however, that the identity for this warp is now at the initialized parameters $\paramO$.

  While one \textit{could} extract an inverse compositional formulation from (\ref{eq:newtemp}), it would require the creation of $NF$ new templates (one for each point-frame combination). To avoid this we warp back to original reference frame according to $\paramO$ where we take the equivalent gradients and inverse compositional update. We compose this final warp as an update to the initialized parameters
  \begin{align}
    \phi(\pt; \Delta \param) &= \W( \W'(\ptt; \Delta \param \boxplus \paramO); \paramO)^{-1} \\
                            &= \W( \W(\pt; \Delta \param \boxplus \paramO); \paramO)^{-1} . \label{eq:phi_a} 
  \end{align}
  To minimize interruption to the flow of this section, we expand the full definition of $\phi$ in Section~\ref{ssec:phi} and show that its partial derivatives with respect to the parameters are now well defined in Section~\ref{ssec:phi_partial}.


  The inverse compositional formulation of (\ref{eq:newtemp}) is then
  \begin{align}
    \sum_{\ptt \in \mathcal{Q}} \| I_0(\phi(\pt; \Delta \param)) - I(\W(\pt; \param)) \|_\gamma
  \end{align}
  which we minimize with respect to $\Delta \param$. The first order Taylor expansion gives
  \begin{align}
    \sum_{\ptt \in \mathcal{Q}} \left\|
      I_0(\phi(\pt; \bm{0}))
      + \nabla I_0 \frac{\partial \phi}{\partial \param} \Delta \param
      - I(\W(\pt; \param))
    \right\|_\gamma \label{eq:phi_ic}
  \end{align}
  where $\nabla I_0$ is assessed at $\pt$, and $\frac{\partial \phi}{\partial \param}$ is assessed at $\bm{0}$.

\subsection{Derivation of the Proxy Warp} \label{ssec:phi}
  Mathematically speaking, the power of the proxy warp $\phi$ comes from shifting the evaluation point of the warp identity from $\bm{0}$ to some initialization $\paramO$. The nature of projective geometry allows us to determine two forms for $\phi$: (\ref{eq:phi_grad}) is better for extracting analytical partial derivatives; and (\ref{eq:phi_update}) is better for extracting analytical warp updates.

  Following from (\ref{eq:phi_a})
  \begin{align}
    \phi(\pt; \Delta \param) &= W( W(\pt; \Delta \param \boxplus \paramO); \paramO)^{-1} \\
    &= \left< \mathbf{R}_0^\T \left( \bar{z}_0 \tilde{\W}(\pt; \param) - \mathbf{t}_0 \right) \right> \\
    &= \left< \mathbf{R}_0^\T \left( \bar{z}_0 \frac{\mathbf{R}' \pt + d' \mathbf{t}'}{[\mathbf{R}' \pt + d' \mathbf{t}']_z} - \mathbf{t}_0 \right) \right> \label{eq:phi0}
  \end{align}
  where $\mathbf{R}'$, $\mathbf{t}'$, and $d'$ are incremented parameters of $\paramO \boxplus \Delta \param$, and $\bar{z}_0$ is the depth in the proxy image as projected by the initial parameters.

\qsection{For Partial Derivatives.} We rearrange (\ref{eq:phi0}) in order to separate the variables and constants
  \begin{align}
    \phi(\pt; \Delta \param) &= \left< \mathbf{R}_0^\T \left( \bar{z}_0 (\mathbf{R}' \pt + d' \mathbf{t}') - \mathbf{t}_0 [\mathbf{R}' \pt + d' \mathbf{t}']_z \right) \right> \nonumber \\
    &= \left< \mathbf{R}_0^\T \left( \bar{z}_0 \mathbf{I} - [\bm{0} ~ \bm{0} ~ \mathbf{t}_0] \right) (\mathbf{R}' \pt + d' \mathbf{t}') \right> \label{eq:phi_grad}.
  \end{align}

\qsection{For Warp Updates.} We multiply the projection by $\frac{1}{d'}$ so that the denominator represents the depth in the proxy image as projected by the updated parameters. For small updates we approximate $\bar{z}_0 [\tfrac{1}{d'} \mathbf{R}' \pt + \mathbf{t}']_z^{-1} \approx 1$:
  \begin{align}
    \phi(\pt; \Delta \param) &= \left< \mathbf{R}_0^\T \left( \bar{z}_0 \frac{\tfrac{1}{d'} \mathbf{R}' \pt + \mathbf{t}'}{[\tfrac{1}{d'} \mathbf{R}' \pt + \mathbf{t}']_z} - \mathbf{t}_0 \right) \right> \\
    &= \left< \mathbf{R}_0^\T \left( \tfrac{1}{d'} \mathbf{R}' \pt + \mathbf{t}' - \mathbf{t}_0 \right) \right> \\
    &= \left< \mathbf{R}_0^\T \left( \tfrac{1}{d'} \mathbf{R}' \pt + \Delta \mathbf{t} \right) \right> \label{eq:phi_update}
  \end{align}

\subsection{Partial Derivatives of Proxy Warp} \label{ssec:phi_partial}
  We note that from the quotient rule we can calculate the derivative of $\phi$ from the derivative of its internal transform by
  \begin{align}
    \frac{\partial}{\partial \param} \left< \mathbf{v}(\param) \right>
    &= \frac{1}{v_z^2}
    \begin{bmatrix}
    \tfrac{\partial v_x}{\partial \param} v_z - v_x \tfrac{\partial v_z}{\partial \param} \\
    \tfrac{\partial v_y}{\partial \param} v_z - v_y \tfrac{\partial v_z}{\partial \param}
    \end{bmatrix}
  \end{align}

  \newcommand{\dR}{(\Delta \mathbf{R}) \mathbf{R}_0}
  \newcommand{\dt}{(\mathbf{t}_0 + \Delta \mathbf{t})}
  \newcommand{\Mvec}{\mathbf{m}}

  Abbreviating the constant portion of (\ref{eq:phi_grad}) as
    \begin{align}
      \Mmat = \mathbf{R}_0^\T \left( \bar{z}_0 \mathbf{I} - [\bm{0} ~ \bm{0} ~ \mathbf{t}_0] \right) ,
    \end{align}
  we define the internal transform in (\ref{eq:phi_grad}) as
    \begin{align}
      \phi^*(\pt; \Delta \param) = \Mmat (\dR \pt + (d_0 + \Delta d) \dt).
    \end{align}

  Having shown in Section~\ref{ssec:ic_ord} that the partial derivative of the original warp $W$ with respect to the inverse depth is non-zero, we are most concerned with showing that it is non-zero for $\phi$. First, we take the partial derivative of $\phi^*$:
    \begin{align}
      \frac{\partial \phi^*}{\partial \Delta d} \bigg\rvert_{\Delta \param = \bm{0}} = \Mmat \mathbf{t}_0 =
      \begin{bmatrix}
      \Mvec_x^\T \\ \Mvec_y^\T \\ \Mvec_z^\T
      \end{bmatrix}
      \mathbf{t}_0
    \end{align}

  Therefore, the partial derivative of $\phi$ with respect to a change in the inverse depth is
    \begin{align}
      \frac{\partial \phi}{\partial \Delta d} \bigg\rvert_{\Delta \param = \bm{0}}
      &= \frac{1}{(\phi^*_z)^2}
      \begin{bmatrix}
      (\Mvec_x \cdot \mathbf{t}_0) \phi^*_z - \phi^*_x (\Mvec_z \cdot \mathbf{t}_0) \\
      (\Mvec_y \cdot \mathbf{t}_0) \phi^*_z - \phi^*_y (\Mvec_z \cdot \mathbf{t}_0)
      \end{bmatrix}
    \end{align}
    which is non-zero for a non-zero initial translation. We only document this process for the inverse depths as the partial derivative of $W$ with respect to the pose parameters are already non-zero; the pre-multiplication of $\Mmat$ does not affect this.

\subsection{Solving the System}
  We solve the system using Gauss-Newton gradient descent. With the inverse compositional formulation, the Jacobian and Hessian matrices need only be calculated once. There is one residual per frame $f$, per point $n$, per patch pixel $\pt$. Each residual and its associated row in the Jacobian is calculated by
  \begin{align}
    \mathbf{J}_{nf\pt} &= \nabla I_0 \frac{\partial \phi(\pt; \Delta \param_{fn})}{\partial \Delta \param_{fn}} \bigg\rvert_{\Delta \param=\bm{0}} \\ 
    \mathbf{r}_{nf\pt} &= I_0(\pt) - I_f(\W(\pt; \param_{fn}))
  \end{align}

  The Hessian is computed from the Jacobian and initial Huber weights by
  \begin{align}
    \mathbf{H} &= \mathbf{J}^\T \mathbf{W}_0 \mathbf{J}.
  \end{align}

  With the Hessian constant under changes to $\param$, a solution to the system is found by iterating the following steps until convergence
  \begin{enumerate}
    \item Recompute weighting matrix $\mathbf{W}$ for Huber loss;
    \item Compute $\mathbf{g} := \mathbf{J}^\T \mathbf{W} \mathbf{r}$;
    \item Compute the inverse update for all poses and inverse depths $\Delta \param := -\mathbf{H}^{-1} \mathbf{g}$;
    \item Update each the warp \\ $W(\pt; \param_{fn}) := W(\phi(\pt; \Delta \param_{fn})^{-1}; \param_{fn})$
  \end{enumerate}


\subsection{Composing the Incremental Warps.}
  In the final step of each iteration, our goal is to approximate the new 6-DoF camera poses and inverse point depths under the updated warps $\W(\phi(\pt; \Delta \param_{fn})^{-1}; \param_{fn})$.
  
  In a similar fashion to Baker and Matthews~\cite{baker2001equivalence}, we approximate the inverse of $\phi$ to the first order in $\Delta \param$ by
  \begin{align}
    \phi(\pt; \Delta \param)^{-1} = \phi(\pt; -\Delta \param)
  \end{align}


  Each warp update is therefore composed as
  \begin{align}
    \W(\phi(\pt; -\Delta \param_{fn}); \param_{fn})
    &= \left< \frac{1}{d} \mathbf{R} \frac{\phi(\pt; -\Delta \param_{fn})}{\phi_z} + \mathbf{t} \right> \label{eq:update1}
  \end{align}
  where $\phi_z$ is the depth of the point under the transform of $\phi$ before it is finally projected into the image, such that $\frac{\phi(\cdot)}{\phi_z}$ is homogeneous.

  We note that the new poses and inverse depths are not strictly independent in from each other in this set of warps. We therefore make minimal approximations such that we may derive analytical and independent updates. Under small deltas we approximate $\phi_z \approx \frac{1}{d_0}$, and substituting (\ref{eq:phi_update}) into (\ref{eq:update1}) we have
  \begin{align}
    \left<
      \underbrace{ \frac{d_0}{d' d } }_{d_{k+1}}
      \underbrace{\vphantom{\frac{1}{1}} \mathbf{R} \mathbf{R}_0^\T \mathbf{R}'}_{\mathbf{R}_{k+1}} \pt +
      \underbrace{\frac{d_0}{d} \mathbf{R} \mathbf{R}_0^\T \mathbf{t}' -%
      \frac{d_0}{d} \mathbf{R} \mathbf{R}_0^\T \mathbf{t}_0 +%
      \mathbf{t}}_{\mathbf{t}_{k+1}} \right> 
  \end{align}
  where $\mathbf{R}' = (\Delta \mathbf{R}^\T) \mathbf{R}$, $\mathbf{t}' = \mathbf{t} - \Delta \mathbf{t}$, and $d' = d - \Delta d$.

  In this canonical form we can read off the new values for the parameters that approximate the updated warp. Approximating $\frac{d_0}{d} \approx 1$ for the translation parameter we calculate the parameters at iteration $k + 1$ as
  \begin{align}
    \mathbf{R}_{k+1} &:= \mathbf{R}_k \mathbf{R}_0^\T (\Delta \mathbf{R}^\T) \mathbf{R}_0 \\
    \mathbf{t}_{k+1} &:= \mathbf{t}_k - \mathbf{R}_k \mathbf{R}_0^\T (\Delta \mathbf{t}) \\
    d_{k+1} &:= \frac{(d_0 - \Delta d)}{d_0} d_k
  \end{align}


\section{Experiments}
  In our experiments we compare the speed and accuracy of the inverse compositional and forwards compositional formulations for photometric bundle adjustment (PBA). In Section~\ref{ssec:sfm} we compare our own ground-up implementations, and in Secton~\ref{ssec:dso} we show our formulation is readily integrated with an existing PBA method (DSO SLAM~\cite{engel2016direct}).

\qsection{Hardware.}
  All experiments are run on a laptop machine with a dual core 2012 Core i5 Intel\textsuperscript{TM} CPU, 8Gb RAM.


\subsection{Structure from Motion} \label{ssec:sfm}
  We compare the speed and equivalence of the inverse versus the forwards compositional formulations on three dataset: (i) The Stanford Bunny sequence from the Stanford Light Field Archive~\cite{vaish2008new} (chosen because it contains a large number of views of the lovable Stanford Bunny); (ii) our own high frame rate recording of handheld camera orbitting an object; and (iii) .

  To compare the forwards and inverse compositional methods we apply Gaussian noise to the base parameters and make note of convergence behaviors of the two methods in terms of the residual and RMS error.

  Unless otherwise specified, we first obtain a base solution to the system by
  \begin{enumerate}
    \item Detect 2000 good Harris corners~\cite{harris1988combined} for tracking;
    \item Robust interest point tracking to obtain correspondences, based on the KLT algorithm~\cite{lucas1981iterative};
    \item Geometric bundle adjustment with outlier removal to obtain initial point depths and camera poses;
    \item Forwards compositional photometric bundle adjustment to refine parameters.
  \end{enumerate}

  Convergence is reached when the largest update of all the reprojected point coordinates is less than a $5\times10^{-3}$ pixels.

  \qsection{Magnitude of Noise.}
  All sequences normalize the mean of the depths to be one. As such increments of a similar magnitude in translation and the angle-axis representation of rotation have a similar effect on the magnitude of the reprojected point displacement.

\subsubsection{Stanford Bunny}
  The 289 of views in the Standford Bunny dataset are collected by moving a camera on an XY gantry and taking photos of the scene at locations on a grid. While the dataset provides ground truth of the gantry position, there is no ground truth of the precise camera intrinsics or extrinsics, and so we obtain the base parameters as described in Section~\ref{ssec:sfm}. Additionally, as a preprocessing step, we downsample the images by a factor of four primarily due to memory limitations.

\begin{figure}[t]
  \centering
  \begin{subfigure}[b]{\linewidth}
    \includegraphics[width=0.9\linewidth, trim=0 -2cm 0 1.25cm, clip=true]{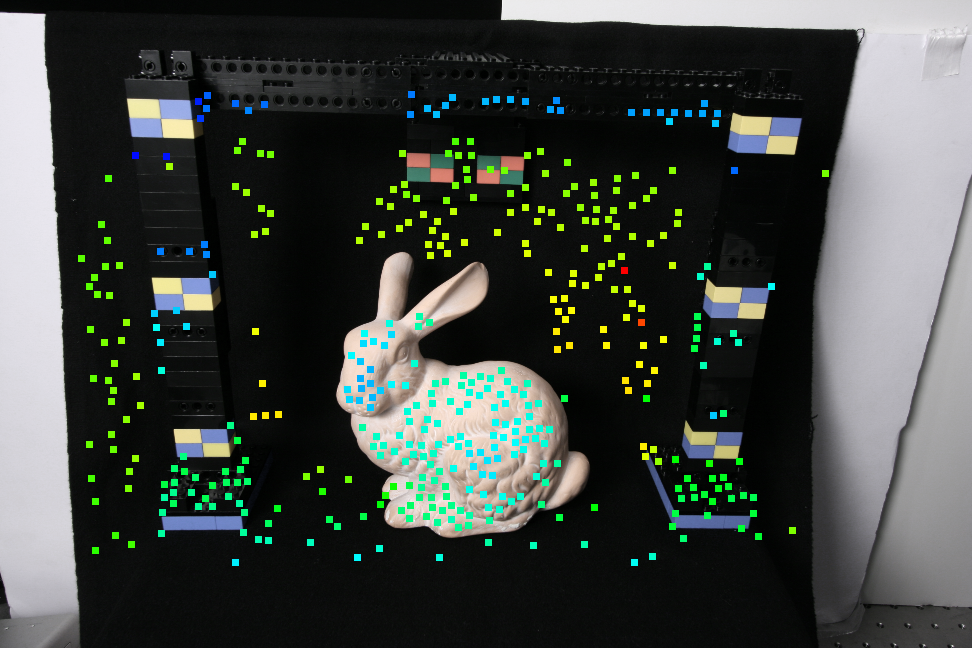}
    \caption{The Standford Bunny sequence from the Stanford Light Light Archive~\cite{vaish2008new} contains 289 views taken by a DSLR on a robotic gantry ((video in supplementary material). The optimization is initialized with 433 points, visualized here by their inverse depths (blue is closer, red is further).}
  \end{subfigure}
  \begin{subfigure}[b]{\linewidth}
    \includegraphics[width=0.9\linewidth, trim=0 -0.4cm 0 0]{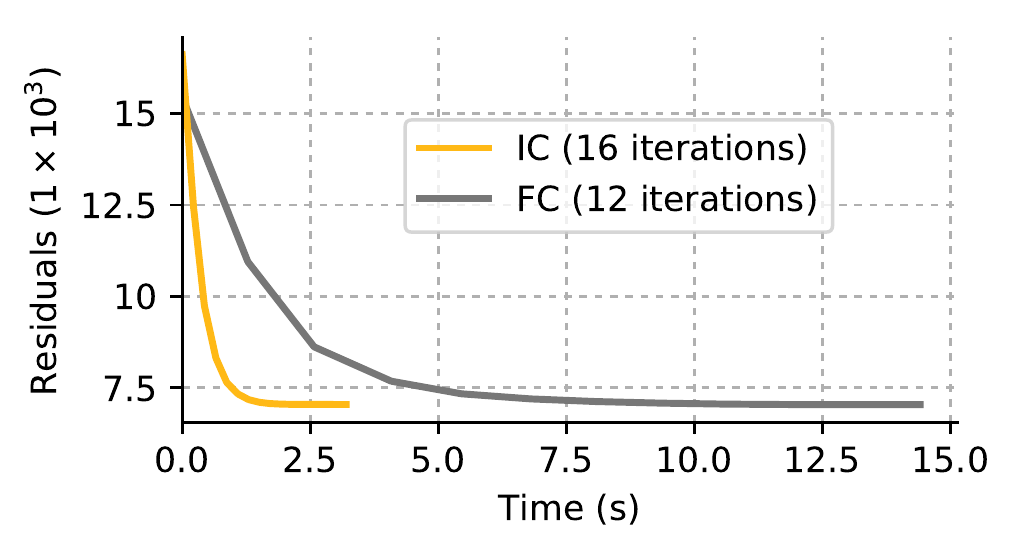}
    \caption{The energy of the system in terms of the intensity residuals.}
  \end{subfigure}
  \begin{subfigure}[b]{\linewidth}
    \includegraphics[width=0.9\linewidth]{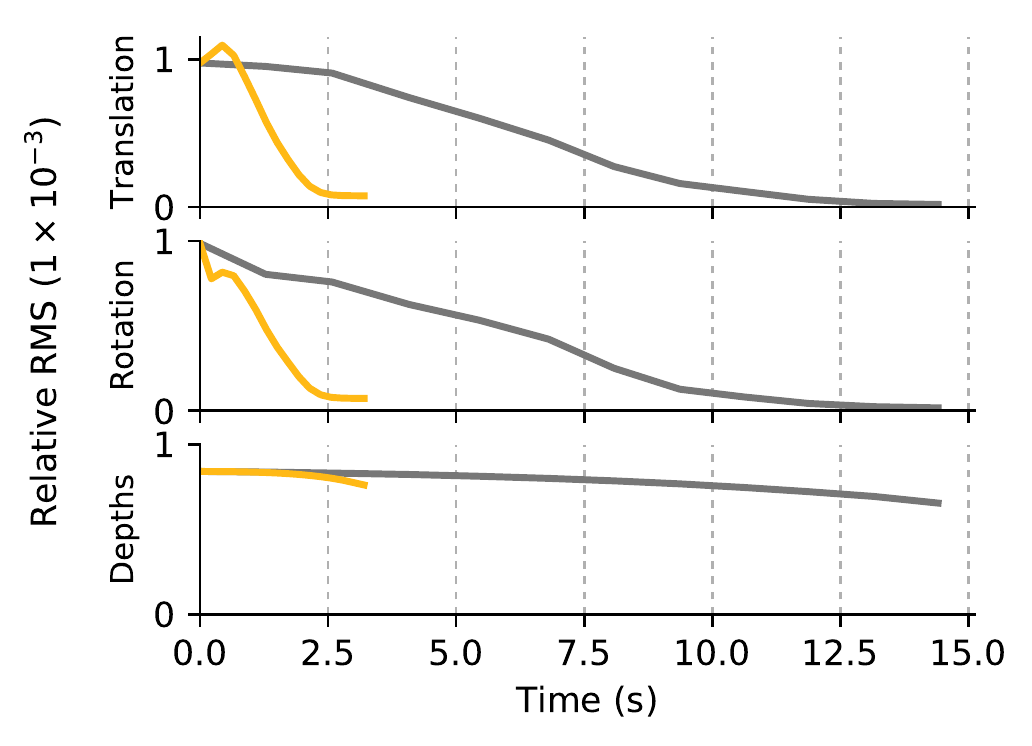}
    \caption{The RMS errors in the parameters relative to the known refined parameters. The mean of the point depths is $1$.}
  \end{subfigure}
    \caption{The inverse compositional (IC) method offers a $4$ times speed-up compared to the forwards compositional (FC) formulation.}
  \label{fig:converge_bunny}
  \end{figure}

  Given the previously refined poses (289 frames) and inverse depths (433 points), we apply Gaussian noise with $\sigma=1\times10^{-3}$ to all parameters before optimizing. We observe in Fig.~\ref{fig:converge_bunny} that while it takes more iterations for the inverse compositional method to converge, it does so in less than a quarter of the time.


\subsubsection{Toy Robot}
\begin{figure}[t]
  \centering
  \begin{subfigure}[b]{\linewidth}
    \includegraphics[width=0.9\linewidth, trim=0 0 0 0, clip=true]{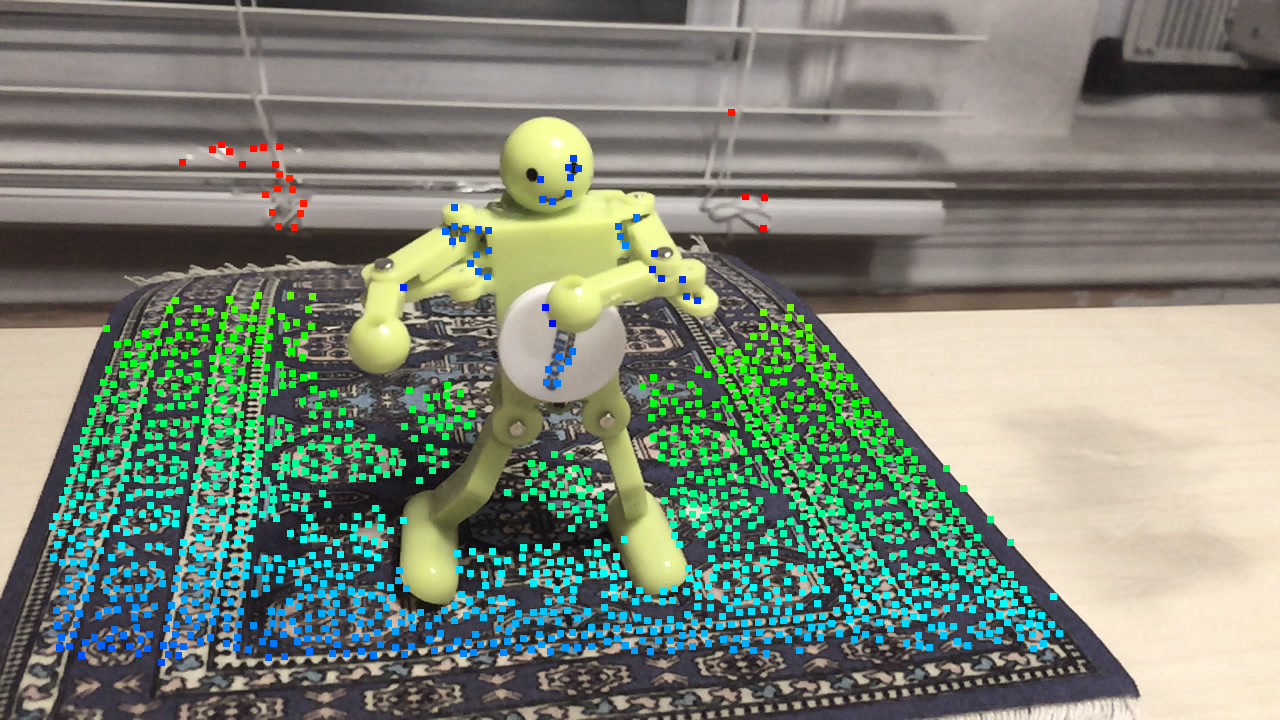} 
    \caption{The Toy Robot sequence contains 300 views captured at 240FPS on an iPhone 6 at $1280\times720$ (video in supplementary material). The optimization is initialized with 1465 points, visualized here by their inverse depths (blue is closer, red is further).}
  \end{subfigure}
  \begin{subfigure}[b]{\linewidth}
    \includegraphics[width=0.9\linewidth, trim=0 0 0 0, clip=true]{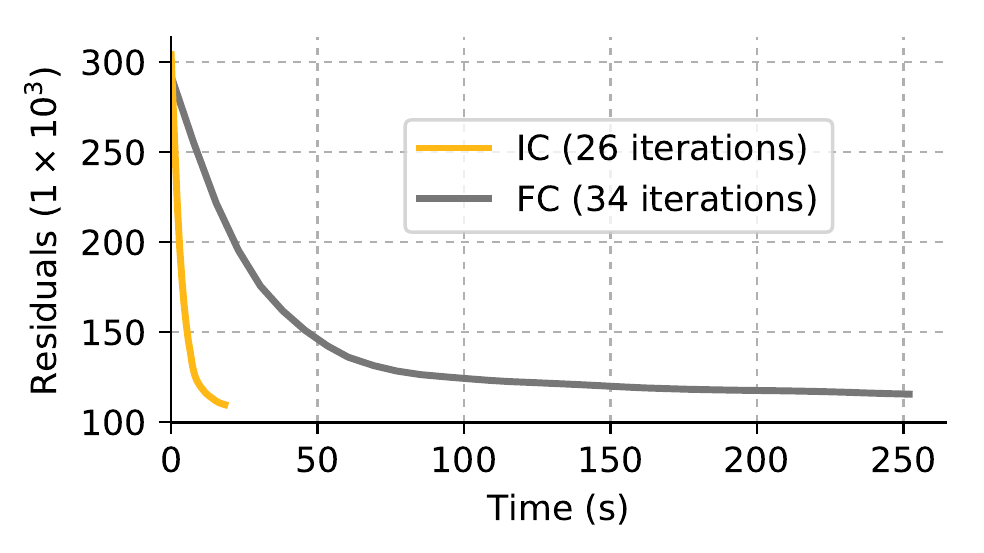}
    \caption{The energy of the system in terms of the intensity residuals.}
  \end{subfigure}
  \begin{subfigure}[b]{\linewidth}
    \includegraphics[width=0.9\linewidth]{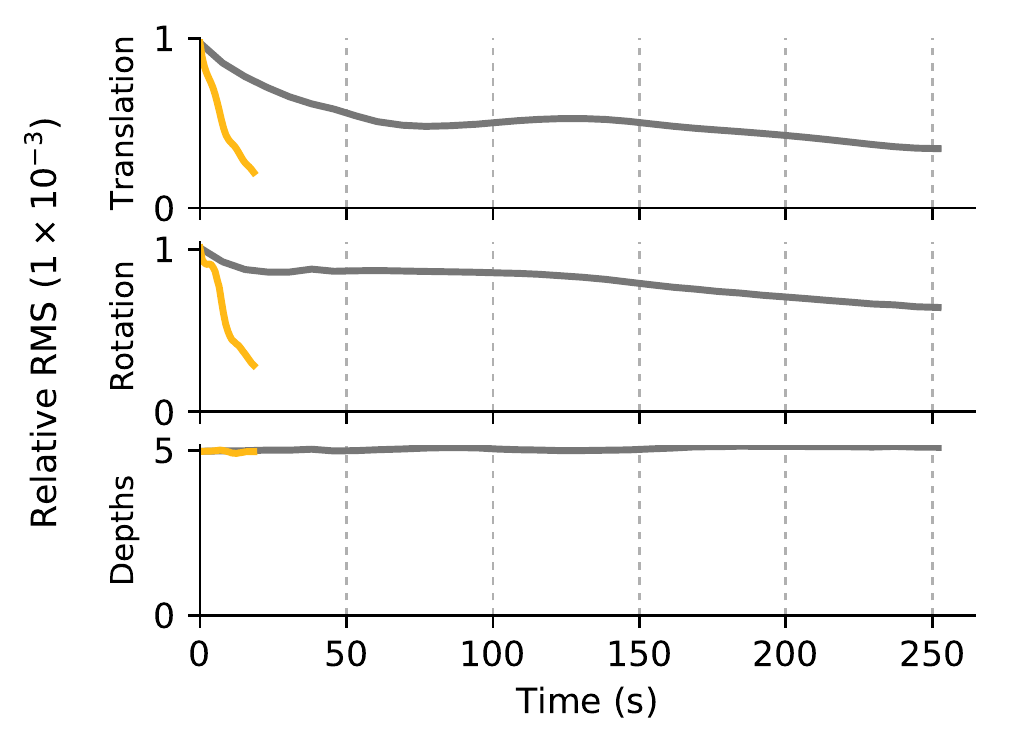}
    \caption{WThe RMS errors in the parameters relative to the known refined parameters. The mean of the point depths is $1$.}
  \end{subfigure}
    \caption{The inverse compositional (IC) method offers a $13$ times speed-up compared to the forwards compositional (FC) formulation.}
  \label{fig:converge_toy}
  \end{figure}

  The toy robot sequence is recorded on an iPhone 6 at 120 FPS with a resolution of $1280\times720$ with image stabalization disabled. Similarly to the Stanford Bunny we apply Gaussian noise with $\sigma=1\times10^{-3}$ to the poses (300 frames) and inverse depths (1465 points). For this sequence the inverse compositional method converges over $13$ times faster, with fewer iterations, and with a lower relative error in the pose estimates (see Fig.~\ref{fig:converge_toy}).


  \begin{figure}[t]
  \centering
  \begin{subfigure}[b]{\linewidth}
    \includegraphics[width=0.9\linewidth]{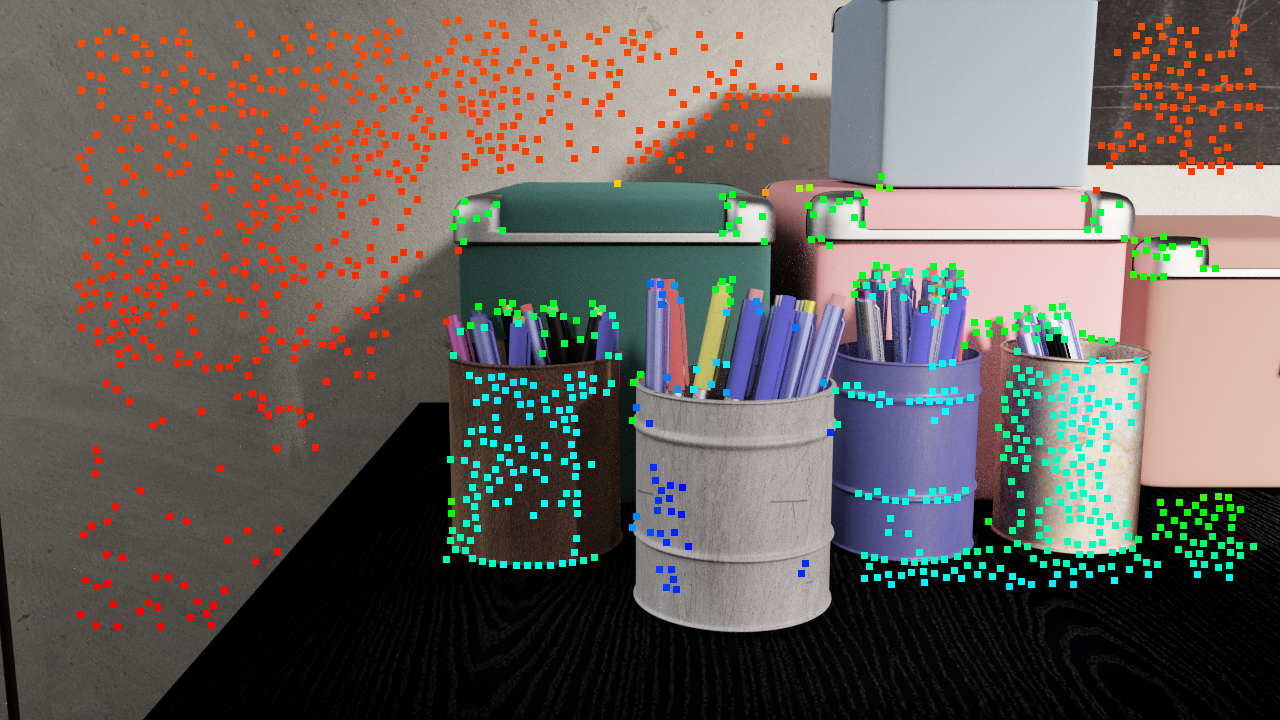}
    \caption{The synthetic sequence contains 100 views rendered at $1280\times720$ (video in supplementary material). The optimization is initialized with 1577 points, visualized here by their inverse depths (blue is closer, red is further).}
  \end{subfigure}
  \begin{subfigure}[b]{\linewidth}
    \includegraphics[width=0.9\linewidth]{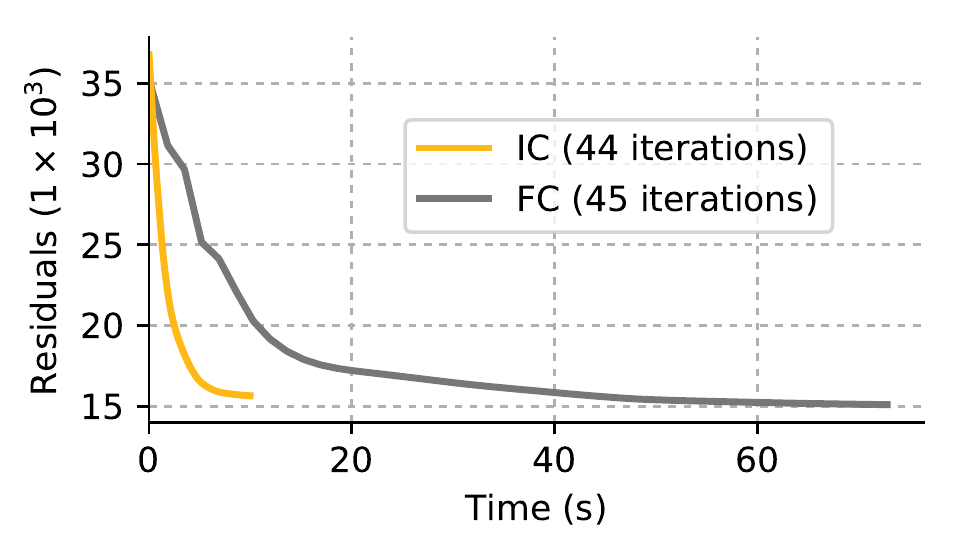}
    \caption{The energy of the system in terms of the intensity residuals}
  \end{subfigure}
  \begin{subfigure}[b]{\linewidth}
    \includegraphics[width=0.9\linewidth]{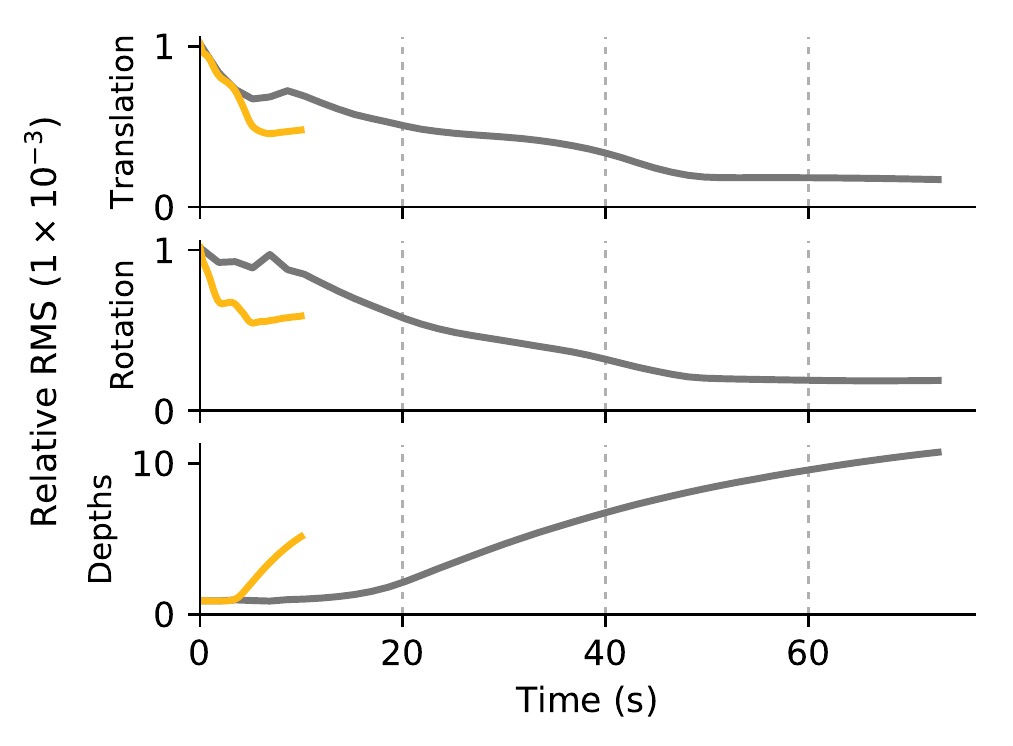}
    \caption{The RMS errors in the parameters compared to the ground truth. The mean of the point depths is $1$.}
  \end{subfigure}
    \caption{The inverse compositional (IC) method offers a $7$ times speed-up compared to the forwards compositional (FC) formulation.}
  \label{fig:converge_synth}
  \end{figure}

\subsubsection{Synthetic Ground Truth}
  In order to be able to compare against ground truth, we also include our own synthetically generated dataset. It is a photorealistically rendered scene using Blender\textsuperscript{TM} for which we have ground truth pose and depth map of the reference frame.

  Figure~\ref{fig:converge_synth} shows the RMS error of the parameters with respect to ground truth and the residual errors. The inverse compositional method is over $7$ times as fast and converges with more error in the pose estimates and less in the inverse depths.


\subsection{Direct Sparse Odometry} \label{ssec:dso}
  To demonstrate that our proposed proxy templates method can enable an inverse compositional formulation for existing methods, we integrate it with DSO~\cite{engel2016direct}. The photometric component of DSO is designed and highly optimized to refined a window of short term keyframes for large scale odometry. Even though in our implementation a new Hessian must still be created whenever a new keyframe is added or dropped, we still observe a $40\%$ increase in the tracking frame rate. We expect that further speed-ups could be found by implementating a dynamic Hessian.

  When running this experiment, DSO was configured to allow a maximum of $30$ frames.

\section{Conclusion}
  In this paper we have shown why an ordinary inverse compositional formulation is not possible for photometric bundle adjustments parameterized by 6 DoF camera poses and inverse depths in a reference frame due to ill-defined derivatives at the identity warp. However, we propose a proxy warp that shifts the evaluation point of the identity warp to be at some initial estimation of the parameters, allowing for valid inverse compositional updates to be estimated. We have shown an order of magnitude speed-up by using the inverse in lieu of the forwards compositional method.

  \begin{figure}[h]
  \centering
  \begin{subfigure}[b]{\linewidth}
    \includegraphics[width=0.9\linewidth]{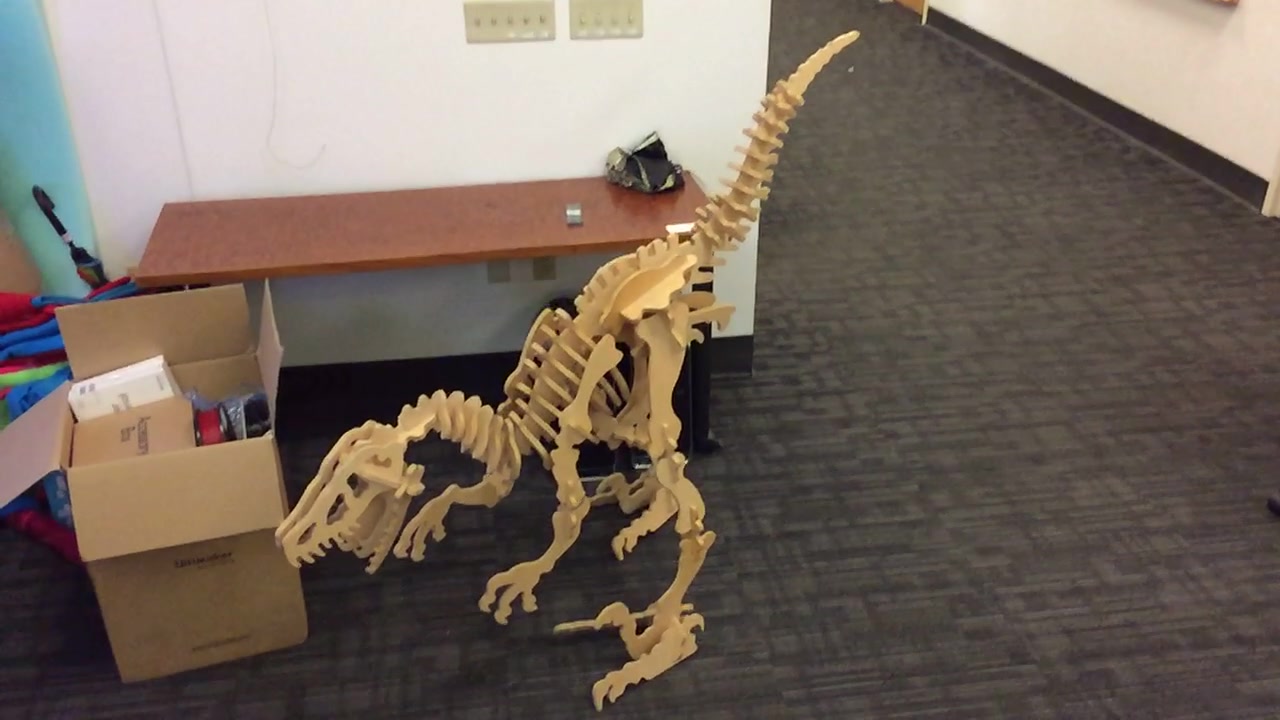}
    \caption{The laser cut dinosaur sequence contains 5423 views captured using the high frame rate (120FPS) setting on an iPhone 6 at $1280\times720$ (video in supplementary material).}
  \end{subfigure}
  \begin{subfigure}[b]{\linewidth}
    \includegraphics[width=1.0\linewidth, trim=14cm 10cm 6cm 8cm, clip=true]{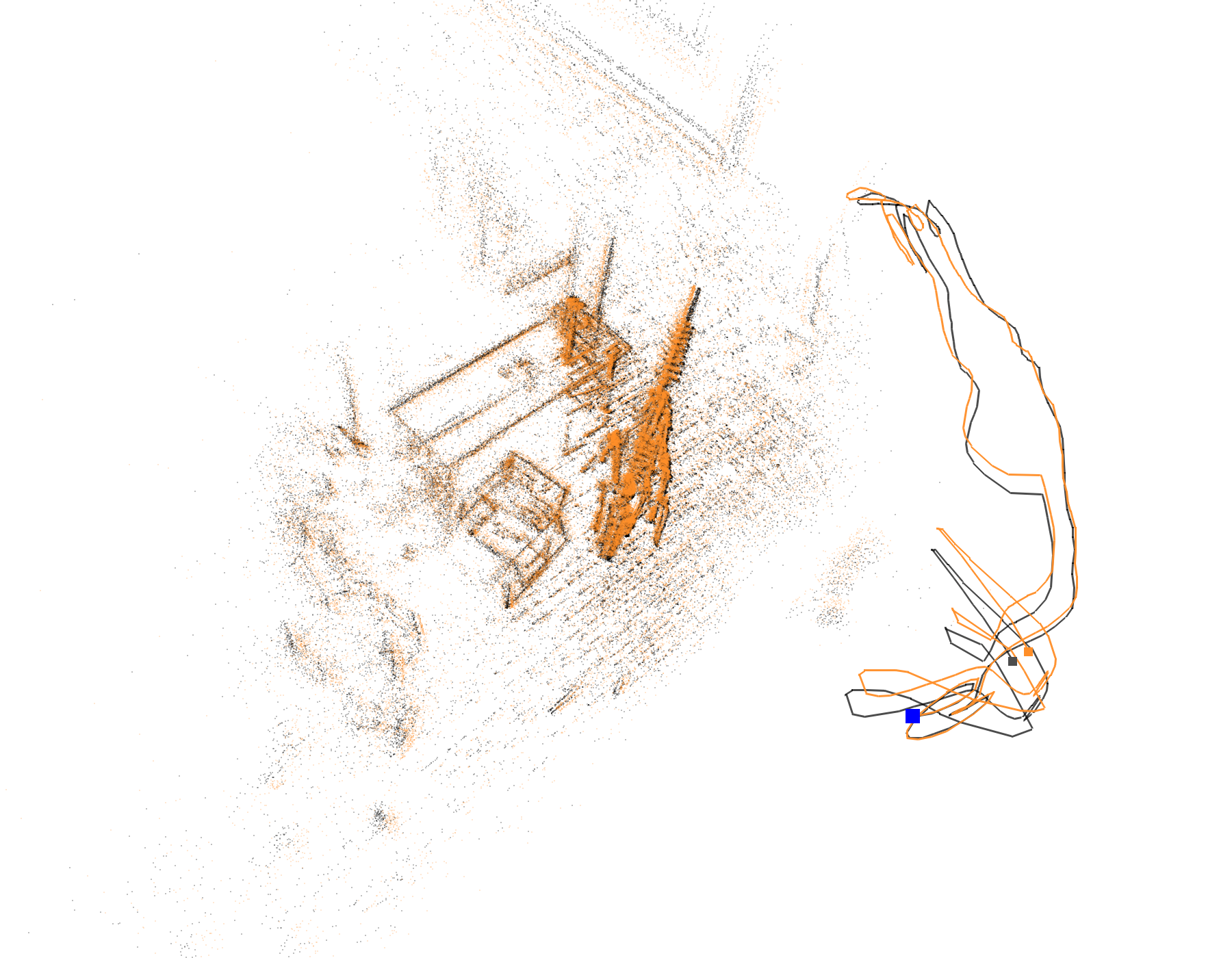}
    \caption{Point cloud and camera paths of the inverse (\textit{orange}) and forwards (\textit{grey}) compositional methods.}
  \end{subfigure}
    \caption{The sequence is tracked using ordinary DSO~\cite{engel2016direct} and DSO modified to optimize using the inverse compositional formulation. IC takes $154$s while FC takes $220$s to track the sequence ($40\%$ increase in frame rate). The distance between the inverse (\textit{orange dot}) and forwards (\textit{grey dot}) camera centers at the end of the sequence is $0.047$. For an idea of scale the furthest camera from the origin (\textit{blue dot}) is $1.39$.}
    \label{fig:dino_dso}
  \end{figure}

{\small
\bibliographystyle{ieee}
\bibliography{ref}
}

\end{document}

%% file: figures/intro.pdf_tex
\begingroup%
  \makeatletter%
  \providecommand\color[2][]{%
    \errmessage{(Inkscape) Color is used for the text in Inkscape, but the package 'color.sty' is not loaded}%
    \renewcommand\color[2][]{}%
  }%
  \providecommand\transparent[1]{%
    \errmessage{(Inkscape) Transparency is used (non-zero) for the text in Inkscape, but the package 'transparent.sty' is not loaded}%
    \renewcommand\transparent[1]{}%
  }%
  \providecommand\rotatebox[2]{#2}%
  \ifx\svgwidth\undefined%
    \setlength{\unitlength}{240.94488189bp}%
    \ifx\svgscale\undefined%
      \relax%
    \else%
      \setlength{\unitlength}{\unitlength * \real{\svgscale}}%
    \fi%
  \else%
    \setlength{\unitlength}{\svgwidth}%
  \fi%
  \global\let\svgwidth\undefined%
  \global\let\svgscale\undefined%
  \makeatother%
  \begin{picture}(1,1.05882353)%
    \put(0,0){\includegraphics[width=\unitlength,page=1]{figures/intro.pdf}}%
    \put(0.23847516,0.99645538){\color[rgb]{0,0,0}\makebox(0,0)[b]{\smash{$\W(\pt; \paramO)$}}}%
    \put(0.78631823,0.99645538){\color[rgb]{0,0,0}\makebox(0,0)[b]{\smash{\warpR}}}%
    \put(0.24511568,0.44861228){\color[rgb]{0,0,0}\makebox(0,0)[b]{\smash{\warpt}}}%
    \put(0.79295875,0.44861228){\color[rgb]{0,0,0}\makebox(0,0)[b]{\smash{\warpd}}}%
    \put(0.24403919,0.78856211){\color[rgb]{0,0,0}\makebox(0,0)[lb]{\smash{$\pt$}}}%
  \end{picture}%
\endgroup%